\documentclass[letterpaper,10pt,journal,twoside]{IEEEtran}

\newcommand{\manuscriptroot}{manuscript/}

\usepackage[table]{xcolor}

\newcommand{\papertitle}{RAIPAL: A Powerful and Agile Humanoid Arm for Human Tasks}
\newcommand{\paperauthors}{Anonymous Authors}

\title{\papertitle}
\author{\paperauthors}

\usepackage{amsmath,amsfonts,amssymb}
\usepackage{array}
\usepackage{textcomp}
\usepackage{stfloats}
\usepackage{url}
\usepackage{verbatim}
\usepackage{cite}

\usepackage{graphicx}
\usepackage{graphics}
\usepackage{multirow}
\usepackage{hyperref}
\usepackage{tabularx}
\usepackage{soul}
\usepackage{makecell}
\usepackage{gensymb}
\usepackage{booktabs}
\usepackage{indentfirst}
\usepackage[normalem]{ulem}
\usepackage{lipsum}
\usepackage{framed}
\usepackage{float}
\usepackage{svg}
\usepackage{pdfpages}

\newcommand{\ph}[1]{{\textbf{#1:}}}
\newcommand{\rph}[1]{{\textbf{}}}

\makeatletter
\renewcommand{\thesubsubsection}{\arabic{subsubsection}}
\def\@seccntformat#1{%
  \@ifundefined{#1@cntformat}%
    {\csname the#1\endcsname\quad}%
    {\csname #1@cntformat\endcsname}%
}
\expandafter\def\csname subsubsection@cntformat\endcsname{\thesubsubsection)\enspace}
\newcommand{\subsubsectionruninformat}[1]{%
  \normalfont\normalsize\bfseries\itshape #1%
}
\renewcommand\subsubsection{%
  \@startsection{subsubsection}{3}{0pt}%
    {0.5ex plus 0.2ex minus 0.1ex}%
    {-0.35em}%
    {\subsubsectionruninformat}%
}
\makeatother

\graphicspath{{\manuscriptroot figures/}}

\usepackage[caption=false,font=footnotesize,position=bottom]{subfig}

\makeatletter
\renewcommand{\p@subfigure}{\thefigure}

\makeatother
\newcommand{\RALacceptedmonthyear}{September, 2026}

\newcommand{\RALfirstauthorsurname}{Kim}
\newcommand{\RALshorttitle}{\papertitle}

\title{\papertitle}

\author{%
Donggeon Kim$^{1}$, Hyungho Chris Choi$^{1}$, Donghyuk Choi$^{1}$, 
Jeongsoo Park$^{1}$, Choongin Lee$^{1}$, Chan Park$^{1}$, and Jemin Hwangbo$^{*, 1}$
\thanks{Manuscript received: May 3, 2026; Revised: June 26, 2026; Accepted: August 27, 2026.}
\thanks{This paper was recommended for publication by Editor Cosimo Della Santina upon evaluation of the Associate Editor and Reviewers' comments.
This work was supported by the Technology Innovation Program(or Industrial Strategic Technology Development Program-Robot Industry Technology Development)(RS-2024-00427719, Dexterous and Agile Humanoid Robots for Industrial Applications) funded By the Ministry of  Trade Industry \& Energy(MOTIE, Korea), and by the National Research Council of Science \& Technology(NST) grant by the Korea government (MSIT) (No. GTL25042-000) \textit{(Donggeon Kim, Hyungho Chris Choi, and Donghyuk Choi contributed equally to this work.) (Corresponding author: Jemin Hwangbo.)}}%

\thanks{$^{1}$Donggeon Kim, Hyungho Chris Choi, Donghyuk Choi, Jeongsoo Park, Choongin Lee, Chan Park, and Jemin Hwangbo are with Robotics and Artificial Intelligence Lab, KAIST, Daejeon, South Korea
        {\tt\footnotesize (email: kcoin@kaist.ac.kr, hyunghochoi@kaist.ac.kr, cay879@kaist.ac.kr, integerpark@kaist.ac.kr, choongin.lee@kaist.ac.kr, pchan0532@kaist.ac.kr, jhwangbo@kaist.ac.kr)}}%
\thanks{Digital Object Identifier (DOI): see top of this page.}%
}

\begin{document}

\bstctlcite{IEEEexample:BSTcontrol}

\maketitle
\begin{abstract}
\label{section:abstract}
This paper presents humanoid manipulator hardware designed specifically for human-level performance throughout its broad domain of tasks, from heavy payloads to highly dynamic motions.
Unlike conventional designs targeting uniform workspace capability, the proposed approach derives
force, speed, and endurance targets from published measurements of task-relevant human upper-limb performance.
These targets guide the actuator, mechanism, and thermal design toward torque transparency and sustained operation.
The platform integrates force-oriented upper-arm components, speed-oriented forearm components, and a posture-dependent elbow mechanism that shifts its torque-speed behavior.
These components are driven by planetary actuators and compact actuator-embedded electronics.
Experimental results validate broad task capability, demonstrating a peak payload of 20 kg (2.25$\times$ its own mass), an endpoint speed of 20.6 m/s, and {continued operation under repeated hammering impacts}.
Thermal analysis further confirms that the robot's sustained operation capability exceeds
the published human load--duration boundary
across {nearly} the {entire} load spectrum.
These findings show that
hardware targets derived from human-task measurements narrow
the human-to-humanoid capability gap.
A supplementary video is available at https://youtu.be/xVUWFICD5G0
 
\end{abstract}


\vspace{-6pt}
\section{Introduction}
\label{section:01_introduction}
\vspace{-3pt}

\rph{Human-Task Gap}

\enlargethispage{-\baselineskip}
\IEEEPARstart{A}{dvances} in permanent-magnet motors and specialized robotic actuators let modern humanoids surpass average human benchmarks in isolated capabilities \cite{IIT_Bi-Manual,RAIAthenaZero2026}.
However, payload and fast motion typically reflect different hardware priorities on different platforms, as shown in Table~\ref{tab:humanoid_benchmarking}.
General-purpose humanoids emphasize payload and runtime, whereas speed-oriented systems remain specialized and impractical for basic payload requirements.
Characteristics of human upper-limb tasks instead reveal a coordinated capability envelope spanning heavy-duty manipulation, rapid distal motion, compliant interaction, and load-dependent
endurance \cite{Winter2009}.
The direct use of these task priors and quantitative boundaries as hardware-design targets, however, remains largely unexplored.

\setlength{\textfloatsep}{6pt plus 1pt minus 2pt}
\setlength{\floatsep}{6pt plus 1pt minus 2pt}
\begin{figure}[!t] %
    \centering
    \captionsetup[subfloat]{position=bottom,farskip=2pt,nearskip=2pt}

    \subfloat[Verified payload of 20\,kg for 60\,s.\label{subfig:demo-composite-payload-20kg}]{
        \includegraphics[width=0.9\columnwidth,height=5cm]{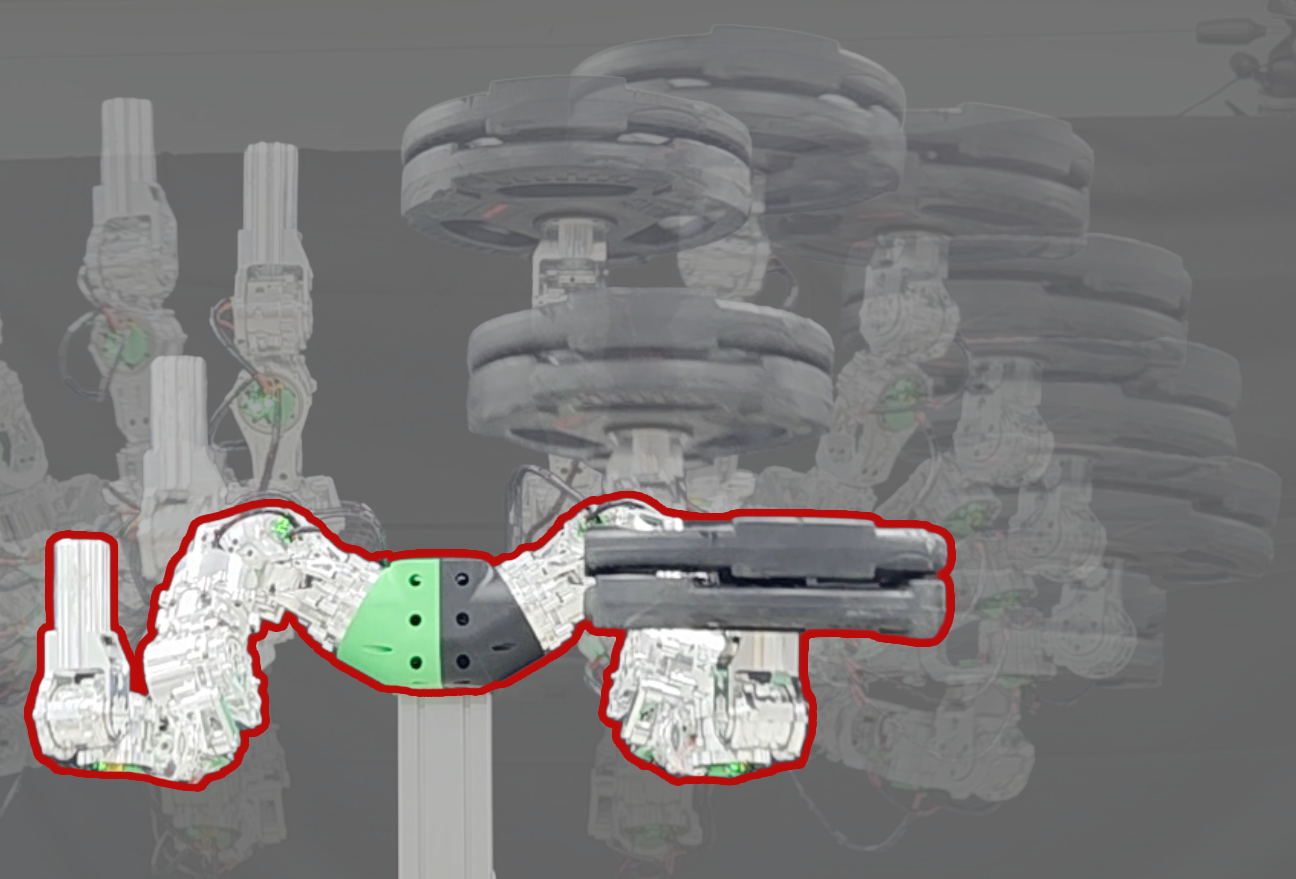}
    }

    \subfloat[Verified endpoint speed of 20.6\,m/s.\label{subfig:demo-composite-speed-20p6mps}]{
        \includegraphics[width=0.9\columnwidth,height=5cm]{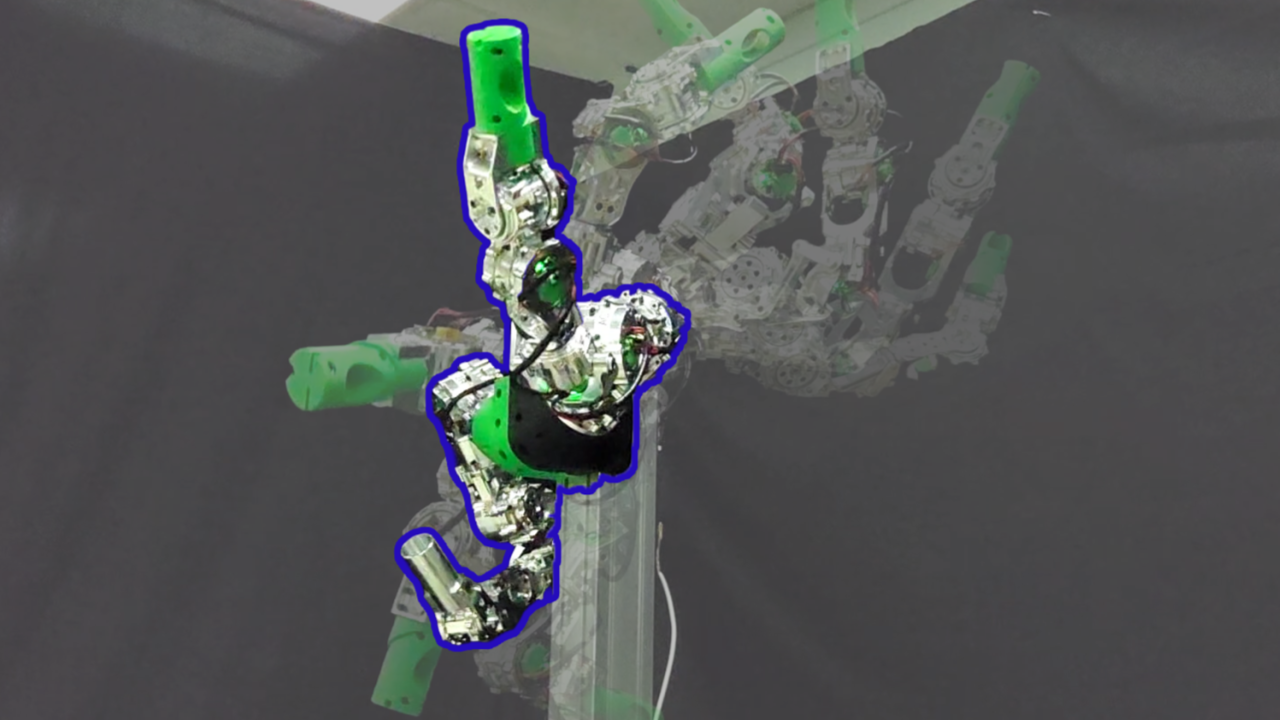}
    }

    \caption{Representative demonstrations. High-payload manipulation and high-speed upper-limb motion achieved with the same humanoid arm.}
    \label{fig:demo_mashup}
\end{figure}

\rph{Design Bias}
Conventional manipulator sizing projects worst-case task-space forces into joint-space torque requirements, enforcing payload capability across broad workspace regions \cite{Sciavicco2000}.
These guarantees suit fixed-base industrial manipulators but poorly match human tasks realized through coordinated, posture-dependent motion.
At the actuator level, the same bias is reinforced by conventional transmission choices: high-reduction precision drives favor compact torque generation but penalize speed, reflected inertia, and torque transparency, whereas low-reduction quasi-direct-drive actuators favor backdrivability and dynamic interaction but require larger motors for heavy-duty tasks proportional to human arms \cite{ARMADA}.
As a result, conventional sizing and actuator tradeoffs tend to push humanoid arms toward either \textit{strong and slow} or \textit{fast and weak} designs, rather than broad human-task capability.
\newpage
\enlargethispage{-2\baselineskip}

\rph{Proposed Work}
This paper proposes a 7-DoF humanoid manipulator built for the wide domain of practical human tasks.
Instead of enforcing uniform workspace-wide capability, the proposed approach
extracts engineering targets from published measurements and analyses of task-relevant human upper-limb force, speed, posture, and endurance.
The resulting prototype, shown in Fig.~\ref{fig:demo_mashup}, is validated through real-world experiments and analysis.
The key contributions of this work are as follows:
\begin{itemize}
    \item A source-grounded method for deriving robot force, speed, endurance, and mass-distribution requirements from measurements of human upper-limb tasks.
    \item A top-down methodology that maps those task-derived parameters into force-, speed-, and posture-specialized hardware components.
    \item Experimental validation through high-payload, high-speed, endurance, and high-impact demonstrations.
\end{itemize}

\begin{table*}[!!!hbt]
    \caption{Comparison of key hardware and task-level capabilities across selected state-of-the-art platforms}
    \vspace{-5pt}
    \label{tab:humanoid_benchmarking}
    \centering
    \footnotesize
    \renewcommand{\arraystretch}{1.1}
    \resizebox{\textwidth}{!}{
        \begin{tabular}{lccccccccc}
            \hline
            Metric 
            & Proposed Arm 
            & {KUKA-DLR LWR}\cite{KUKA_DLR_LWR} 
            & D3-ARM\cite{D3-ARM} 
            & IIT Bi-Manual\cite{IIT_Bi-Manual} 
            & ARMADA\cite{ARMADA} 
            & LIMS\cite{Lims-ambidex} 
            & AthenaZero\cite{RAIAthenaZero2026} 
            & DynaArm\cite{DuaticDynaArm} \\
            \hline
            
            DoF (one arm) 
            & 7 & {7} & 6 & 7 & 6 & 7 & 7 %
            & 6 \\
            
            Mass (one arm) 
            & 8.9\,kg 
            & {15.0\,kg} 
            & 16.5\,kg 
            & 9.0\,kg 
            & 1.09\,kg$^{\ast}$ 
            & 5.84\,kg 
            & N/A 
            & 9.2\,kg \\
            
            Peak Payload 
            & 20.0\,kg 
            & {N/A} 
            & 2.0\,kg 
            & 15.0\,kg 
            & 2.5\,kg 
            & 3.0\,kg 
            & 3.0\,kg 
            & 12.0\,kg \\
            
            Continuous Payload 
            & 5.0\,kg 
            & {7.0\,kg} 
            & N/A 
            & 7.0\,kg 
            & N/A 
            & N/A 
            & N/A 
            & 6.0\,kg \\
            
            Peak Payload-to-Arm-Mass Ratio
            & 2.25:1 
            & {N/A} 
            & 0.12:1 
            & 1.67:1 
            & 2.29:1$^{\ast}$ 
            & 0.51:1 
            & N/A 
            & 1.30:1 \\
            
            Max Speed 
            & 20.6\,m/s 
            & {N/A} 
            & 1.47\,m/s 
            & N/A 
            & 6.16\,m/s 
            & 5.35\,m/s 
            & 31.0\,m/s$^{\dagger}$ 
            & 10.0\,m/s \\
            
            \hline
        \end{tabular}
    }
    \vspace{-3pt}
    \begin{flushleft}
        \scriptsize
        $^{\ast}$ Reported mass only includes the moving parts, without body-mounted components and the gripper \\
        $^{\dagger}$ Maximum throwing velocity achieved through coordinated torso-waist dynamic trajectory optimization\cite{RAIAthenaZero2026} \\
        \textit{Scope:} This comparison includes peer-reviewed systems and commercial platforms with public video demonstrations \\
        \textit{N/A} indicates that the corresponding specification was not reported in the cited source
    \end{flushleft}
    \vspace{-10pt}
\end{table*}

\section{Methods}
\label{section:02_methods}
\subsection{The Human Prior}
\label{subsec:human_task_characteristics}

\rph{Motivation}
Published evidence and measurements of human upper-limb tasks provide the task-relevant engineering parameters used to design
this robot platform.
These task characteristics span high-force and high-speed motion,
mass distribution, posture-dependent behavior,
compliant interaction, and load-dependent endurance, which together
define the intended domain of a high-performing humanoid arm.
Rather than reproducing human anatomy, these task characteristics are translated into
robot design {requirements} such as 
endpoint force, joint speed, torque transparency, and sustained operation.

\ph{Feedback-Driven Accuracy}
Characteristics of contact-rich human tasks indicate that
safe and accurate interaction
combines mechanical compliance with
continuous feedback rather than
relying on extreme stiffness
\cite{Winter2009}.
This compliance enables the arm to yield under contact while maintaining precision through closed-loop correction.
{Consequently, the mechanical design should prioritize torque transparency and low drivetrain impedance so that feedback-based interaction control is not hindered by excessive mechanical resistance.}

\ph{Upper-Arm Strength}
Characteristics of high-force
upper-body tasks contract the arm posture to shorten the external moment arm
about the proximal joints
\cite{Faity2021TorqueReserve}.
This posture uses joint constraints to support large forces through the upper-arm.
The forearm remains nearly parallel to the force, transmitting proximal force while requiring mainly stabilizing torque.

\ph{Forearm Acceleration}
Characteristics of high-velocity
upper-limb tasks instead
include an extended arm posture that increases
the radius of rotation.
The forearm accelerates its low-inertia distal segments to generate wrist-snap motion \cite{Hirashima2008KineticChain}.

\ph{Elbow Mediation}
The elbow mediates between the upper-arm and forearm, and must support both high-torque and high-velocity tasks.
The contracted high-torque and extended high-speed patterns require the elbow to provide high torque in flexion and high velocity in extension.
In the human elbow, the interplay of moment arms and muscle coordination realizes this behavior, producing larger torque when flexed and higher velocity when extended \cite{An1981}.

\ph{Endurance}
Published sustained-exertion measurements show that human upper-limb task endurance
varies strongly with applied load.
For humanoid arms, this
task-derived load--duration boundary
translates into a thermal requirement because heat accumulation in the actuators and embedded electronics limits the duration of sustained joint torque.
Rohmert's maximum-endurance-time relation is relevant because it provides a normalized human-task endurance boundary that can be compared directly with the actuator thermal Safe Operating Area (SOA) \cite{rohmert1973problems}.
The robot thermal SOA should therefore support sustained operation over the relevant load spectrum.

\vspace{-4pt}
\subsection[Human-Task-Derived Engineering Parameters]
{Human-Task-Derived Engineering Parameters}
\label{subsec:task_derived_parameters}

\rph{Motivation}
The numerical targets in Table~\ref{tab:primary_objective_comparison} are representative rounded adult-male design values rather than physiological limits.
{Each task-specific target uses the reported cohort mean plus one standard deviation (\(\mu+\sigma\)), corresponding to approximately the upper \(16\%\) under a normal-distribution approximation; this defines a concrete above-average benchmark.}
Reported measurements are mapped and rounded to the corresponding
robot axes used in Table~\ref{tab:primary_objective_comparison}.
The maximum-torque inputs are taken directly from axis-specific shoulder, elbow, wrist flexion--extension and deviation, and forearm-rotation measurements
\cite{Lategan2011IsokineticNorms,LateganKruger2007Handgrip,Delp1996WristMoments}.
The peak-speed inputs are mapped and rounded from reported maximum upper-limb joint velocities in sports actions %
\cite{JessopPain2016MaximumVelocities}.
{The \(1.80~\mathrm{m}\), \(80~\mathrm{kg}\) human model provides the common basis for converting reported segment proportions to link lengths and masses and body-mass-normalized torque measurements to peak joint torques, with the resulting model supplying the gravity terms}
\cite{Winter2009}.

\ph{Endpoint Force}
Conventional worst-case task-space forces can be projected to joint-space through Jacobian-transpose and gravity terms as follows \cite{Sciavicco2000}:
\begin{align}
\tau_{\max,i}
\ge
\beta \max_{q}
\left\|
J(q)_{i}^{T} f_{\mathrm{ext}} + g(q)_{i}
\right\|
\label{eq:JT_sizing}
\end{align}
\newpage
where \(q\) is the generalized pose, \(J(q)\) is the Jacobian, \(f_{\mathrm{ext}}\) is the external force objective, \(g(q)\) is the generalized gravity term, and \(\beta\) is a safety factor.
Applying this projection as a uniform workspace constraint can narrow the humanoid-arm task domain, but the projection remains useful for converting task-space objectives into joint-space requirements.
Reversing this projection gives endpoint force, a more relevant heavy-duty metric than direct torque matching, which would oversize proximal links and the elbow actuator.
For each joint, endpoint force is defined as the minimum gravity-compensated end-effector force achievable across configurations:
\begin{align}
f_{\mathrm{end},i}^{H}
:=
\min_{q\in\mathcal Q_H}
\left\|
J_i^H(q)^{-T}
\left(
\tau_{i,\max}^{H}
-
g_i^H(q)
\right)
\right\|
\label{eq:endpoint_force}
\end{align}
where \(\mathcal Q_H\) is the evaluated human configuration set, and \(J_i^H(q)\) and \(g_i^H(q)\) are the Jacobian contribution and gravity term of the anthropometric model.

\rph{Worked Example}
{A worked example is provided for Axis~1. Lategan reports a body-mass-normalized shoulder-extension torque of \(1.23\pm0.21~\mathrm{Nm/kg}\)}
\cite{Lategan2011IsokineticNorms}.
{Applying \(\mu+\sigma\) to the \(80~\mathrm{kg}\) reference gives a maximum joint torque of \(\tau_1^H=(1.23+0.21)(80)=115.2~\mathrm{Nm}\).}
{Using upper-arm, forearm, and hand length fractions of \(0.186H\), \(0.146H\), and \(0.108H\), and mass fractions of \(0.028M\), \(0.016M\), and \(0.006M\), the \(1.80~\mathrm{m}\), \(80~\mathrm{kg}\) model has a \(0.7920~\mathrm{m}\) total reach, a \(4.00~\mathrm{kg}\) total arm mass, and a \(12.804~\mathrm{Nm}\) worst-case gravity torque at full extension}
\cite{KeyserlingArmstrong2008Anthropometry,Winter2009}.
{This yields the endpoint-force projection}
\[
{\displaystyle
f_{\mathrm{end},1}^{H}
=\frac{115.2-12.804}{0.7920}
=129.3~\mathrm{N}
\rightarrow 129~\mathrm{N}.}
\]
{Similar projections are evaluated for each human motion and its corresponding worst-case pose.}

\ph{Peak Velocity}
Agile upper-limb capability is quantified by the maximum achievable angular velocity of each joint.
{The source values are measured peak joint velocities from compatible young-adult male cohorts, and the same \(\mu+\sigma\) selection rule is applied.}
This joint-space metric is appropriate because unlike force requirements, velocity is kinematically determined by \(v = J(q)\dot{q}\), allowing joint-speed targets to transfer directly to task-space speed given comparable topology and link proportions \cite{Sciavicco2000}.
The resulting targets are later compared with real robot parameters in Table~\ref{tab:primary_objective_comparison}.

\rph{Additional Constraints}
Link mass and worst-case angular acceleration are retained as secondary constraints but are not part of Table~\ref{tab:primary_objective_comparison}'s primary endpoint-force and peak-speed comparison.
Robot link masses are kept near \(60\%\) of the corresponding human segment masses, while upper-arm worst-case acceleration is evaluated separately as a secondary design check.

\begin{figure}[!b]
    \centering
    \captionsetup[subfloat]{position=bottom,farskip=2pt,nearskip=2pt}

    \subfloat[Upper-arm actuator.\label{subfig:mhr_exploded}]{%
        \includegraphics[width=0.9\linewidth]{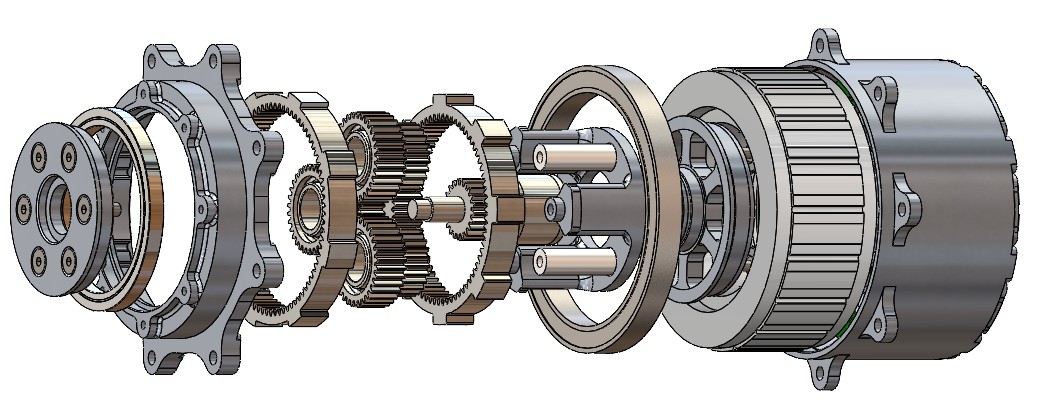}%
    }

    \subfloat[Forearm actuator.\label{subfig:slr_exploded}]{%
        \includegraphics[width=0.9\linewidth]{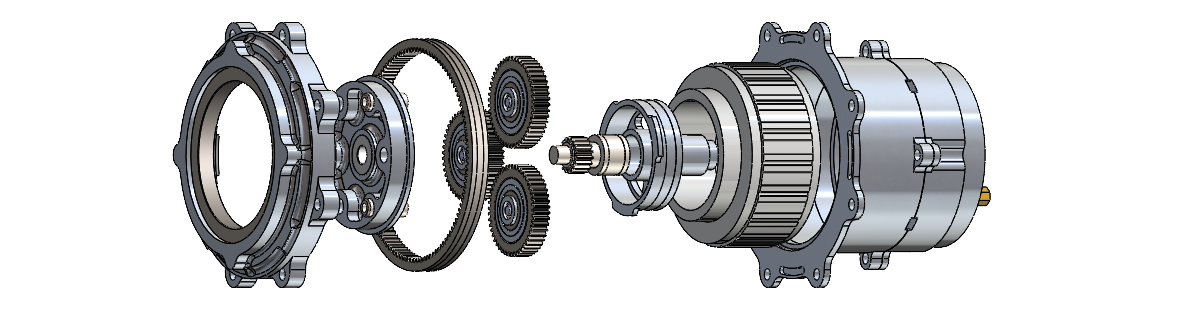}%
    }

    \caption{Exploded actuator families with a 3K compound planetary upper-arm drive and a high-speed, low-inertia planetary forearm drive.}
    \vspace{-5pt}
    \label{fig:actuator_exploded}
\end{figure}

\begin{table}[!b]
    \caption{Specifications of the proposed actuator families}
    \vspace{-5pt}
    \label{tab:actuator_spec}
    \centering
    \footnotesize
    \renewcommand{\arraystretch}{1.1}
    \begin{tabular}{lcc}
        \hline
        Feature & Upper-arm Actuator & Forearm Actuator \\
        \hline
        Mass & 1.50\,kg & 0.36\,kg \\
        Bounding dimensions & $\phi 100 \times 83$\,mm & $\phi 77 \times 60$\,mm \\
        Reduction & 22:1 (3K compound) & 7.33:1 (planetary) \\
        Rated torque & 56\,Nm (30\,min) & 3.88\,Nm (45\,min) \\
        Peak torque & 180\,Nm (26\,s) & 12.61\,Nm (15\,s) \\
        No-load speed (80\,V) & 12.8\,rad/s & 89.7\,rad/s \\
        Reflected inertia & $5.76\times10^{-2}$\,kg$\cdot$m$^{2}$ & $5.41\times10^{-4}$\,kg$\cdot$m$^{2}$ \\
        \hline
    \end{tabular}
    \vspace{-4.5pt}
\end{table}

\newpage
\subsection{Actuator Design}
\label{subsec:actuator_design}

\begin{figure*}[!!b]
    \centering
    \vspace{-3pt}
    \includegraphics[width=\linewidth]{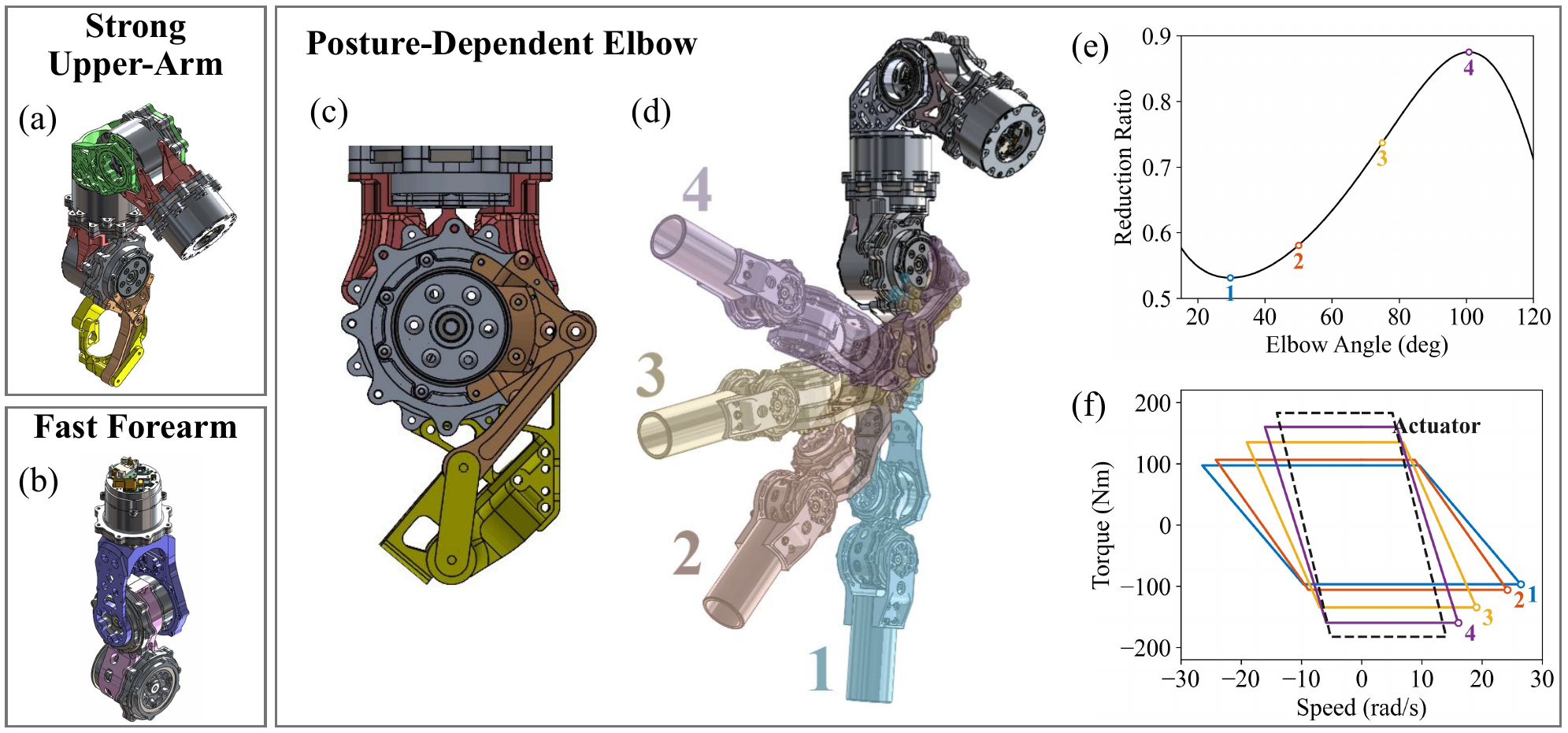}
    \vspace{-6pt}
    \caption{Mechanism-level decomposition of the robot arm design.
    (a) The force-specialized upper-arm and (b) speed-specialized forearm implement the primary segment roles.
    (c,d) The posture-dependent crossed-four-bar elbow changes effective reduction with joint angle.
    (e,f) The resulting transformation expands torque capability in flexed postures and speed capability in extended postures.
    }
    \label{fig:mechanism_overview}
    \vspace{30pt}
\end{figure*}

\rph{Upper-Arm Actuator}
The four upper-arm joints use a 3K{ (also known as a Wolfrom gearbox)} compound planetary transmission {consisting of two axially aligned ring gears, rigidly connected compound planet gears, and a sun gear }(shown in Fig.~\ref{subfig:mhr_exploded}) to provide torque amplification for high-payload manipulation while preserving practical backdrivability.
The selected 22:1 reduction sits between high-reduction precision drives and low-reduction quasi-direct-drive actuators, avoiding the speed and reflected-inertia penalties of very high-reduction transmissions.
The resulting actuator design, shown in Fig.~\ref{subfig:mhr_exploded}, supports the force-oriented upper-arm notion in Table~\ref{tab:primary_objective_comparison}.

\rph{Backlash and Friction Mitigation}
Gear preload trades backlash against static friction: tighter meshing reduces backlash but increases preload-induced friction.
The final preload was selected empirically instead of through iterative tolerance adjustment.
Section~\ref{subsec:upper_arm_actuator_analysis} evaluates the resulting backlash--friction tradeoff.

\rph{Efficiency Analysis}
Gear-mesh Coulomb friction reduces transmission efficiency and torque transparency.
The selected 3K compound planetary transmission is therefore modeled analytically and evaluated experimentally rather than optimized through tooth profiles that could compromise backlash or packaging \cite{Bilateral_Drive_Gear}.
Although serial two-stage planetary transmissions can be more efficient, the 3K topology integrates the required reduction into a compact humanoid upper-body package.
A simpler stepped compound planetary could omit the second ring gear, but achieving 22:1 through the planet-stage size difference would enlarge the planet gears and overall gearbox envelope.
To quantify transmission performance, forward- and backward-driving efficiencies are defined as
\begin{equation}
\eta_{\mathrm{fwd}} =
\frac{
\left|\tau_{\mathrm{out}} \, \omega_{\mathrm{out}}\right|
}{
\left|\tau_{\mathrm{in}} \, \omega_{\mathrm{in}}\right|
},
\quad
\eta_{\mathrm{bwd}} =
\frac{
\left|\tau_{\mathrm{in}} \, \omega_{\mathrm{in}}\right|
}{
\left|\tau_{\mathrm{out}} \, \omega_{\mathrm{out}}\right|
}
\label{eq:efficiency}
\end{equation}
where \((\tau_{\mathrm{in}}, \omega_{\mathrm{in}})\) and \((\tau_{\mathrm{out}}, \omega_{\mathrm{out}})\) denote the input- and output-side torque and angular velocity, respectively.

\rph{Forearm Actuator}
The forearm actuator in Fig.~\ref{subfig:slr_exploded} prioritizes low reflected inertia, high speed, and backdrivability while meeting the distal torque targets from Section~\ref{subsec:task_derived_parameters}.
It uses a 7.33:1 single-stage planetary gearbox, which provides high joint speed and torque transparency for distal acceleration.
Because single-stage planetary transmissions are standard in proprioceptive robotic actuators \cite{RAIBO2}, dedicated transmission analysis focuses on the less conventional 3K compound upper-arm drive.
\vspace{-5pt}
\subsection{Mechanism Design}
\label{subsec:mechanism_design}

\rph{Overview}
The mechanism is decomposed into a proximal force-specialized assembly, a distal speed-specialized assembly, and a posture-dependent elbow transmission.
This follows from the task-dependent force--speed mediation in Section~\ref{subsec:human_task_characteristics}.
The platform terminates at the wrist mounting interface; articulated-hand design and task-specific parameterization are outside this work's scope.

\rph{Strong Upper-Arm (Proximal Force)}
The upper-arm is designed to prioritize torque while remaining as close as possible to the extracted upper-arm velocity targets.
Four repeated high-torque modules add force capacity without joint-specific hardware complexity.

\rph{Fast Forearm (Distal Speed)}
The forearm uses a serial arrangement of actuators to minimize distal inertia and prioritize speed while preserving the endpoint force required to stabilize heavy tasks.
Although this arrangement may reduce accessibility during constrained-space manipulation compared with dedicated wrist mechanisms \cite{Lims-ambidex}, it favors simplicity and rapid acceleration.
The distal assembly therefore acts as the limb accelerator during snapping and throwing motions.
At the implemented operating point, it exceeds all extracted wrist-speed targets while maintaining the required distal force margin.

\rph{Posture-Dependent Elbow}
The elbow implements the dual role described in Section~\ref{subsec:human_task_characteristics} using a crossed-four-bar mechanism whose effective reduction varies with angle
\newpage
\begin{align}
r_{\mathrm{eff}}(\theta)
=
{\omega_{\mathrm{in}}(\theta)} \,/\, {\omega_{\mathrm{out}}(\theta)} ,
\label{eq:cfb_ratio}
\end{align}
where \(\theta\) is the elbow angle, and \(\omega_{\mathrm{in}}\) and \(\omega_{\mathrm{out}}\) are the input and output angular velocities.
Similar posture-dependent joint mechanisms have previously been developed to tailor mechanical advantage to configuration-dependent functional requirements \cite{tomishiro2019design, lenzi2011robotic}.
As shown in Fig.~\ref{fig:mechanism_overview}, the implemented mechanism varies \(r_{\mathrm{eff}}\) from \(0.531\) near extension to \(0.875\) near flexion, expanding the output speed region in extended postures and the output torque region in flexed postures.
For {simulation}, a polynomial surrogate of the explicit formulation introduces negligible computational overhead{}.
The linkage also shifts actuator mass proximally, reducing distal inertia and the effective inertia seen at the shoulder joints by 15\%.
The utility of this posture-dependent operating-region transformation is evaluated experimentally in Section~\ref{subsec:demonstrations}.

\rph{Objective Analysis}
Table~\ref{tab:primary_objective_comparison} compares the resulting robot arm design to the human-task-derived engineering targets set in Section~\ref{subsec:task_derived_parameters}.
The single-arm assembly weighs 8.9~kg, below an estimated 12.3~kg functionally matched human reference that includes 9.0~kg of base musculature driving similar motion but excludes the hand and extrinsic digit-actuating forearm musculature %
\cite{Holzbaur2007UpperLimb}.
Despite the reduced mass, the system exceeds the proximal endpoint-force targets and achieves joint maximum speeds beyond the human reference.
The realized shoulder axes provide 248--502~N of endpoint force, whereas the forearm axes provide 78.6--160.7~N while operating at substantially higher speeds, confirming the deliberate assignment of different force and speed objectives to the two arm regions.
\begin{table}[t]
    \caption{Comparison against human-task-derived engineering targets}
    \label{tab:primary_objective_comparison}
    \centering
    \renewcommand{\arraystretch}{1.12} %
    \footnotesize
    \setlength{\tabcolsep}{4.2pt}
    \begin{tabular}{@{}c l cc cc@{}}
        \hline
        Joint Axis & Joint motion & \multicolumn{2}{c}{Endpoint force (N)} & \multicolumn{2}{c}{Max speed (rad/s)} \\
        \cline{3-4} \cline{5-6}
             &              & Target & Robot & Target & Robot\textsuperscript{a} \\
        \hline
        1 & Shoulder flex./ext. & 129 & 287   & 9.0  & 12.8 \\
        2 & Shoulder abd./add.  & 153 & 248   & 8.0  & 12.8 \\
        3 & Shoulder rot.       & 145 & 502   & 12.0 & 12.8 \\
        4 & Elbow flex./ext.    & 171 & 255   & 22.0 & 24.1 \\
        5 & Wrist pron./sup.    & 84  & 89.3  & 23.0 & 89.7 \\
        6 & Wrist flex./ext.    & 79  & 78.6  & 40.0 & 89.7 \\
        7 & Wrist dev.          & 58  & 160.7 & 20.0 & 89.7 \\
        \hline
    \end{tabular}
    
    \vspace{2pt}
    \parbox{\linewidth}{\footnotesize \textsuperscript{a} Calculated at 80 V.
    }
    \vspace{-15pt}
\end{table}

\begin{figure}[!t]
\centering
    \includegraphics[width=\linewidth]{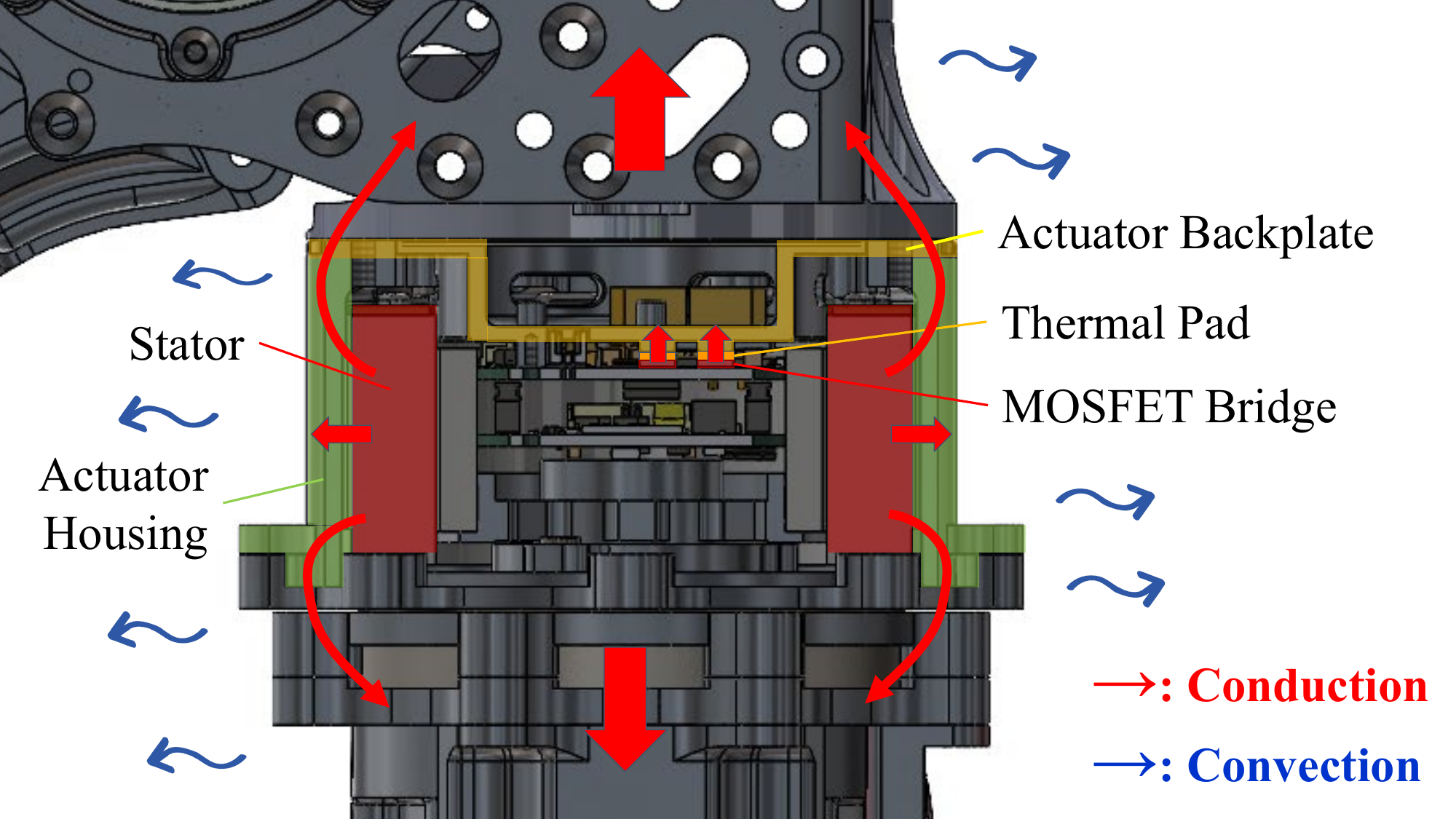}
    \vspace{-15pt}
    \caption{
    Chassis conductive cooling paths of the integrated actuator.
    Red arrows indicate conductive heat flow from the stator and MOSFET bridge into the actuator chassis and attached robot-arm mechanism; blue arrows indicate convection from exposed surfaces.
    }
\label{fig:heat_path}
\end{figure}

\newpage
\subsection{Electronics Design}
\label{subsec:electronics_design}

\rph{Dual-Board Architecture}
Embedding a high-current servo drive inside each actuator places switching electronics close to sensing and communication circuits, making low-voltage signals vulnerable to noise.
A stacked dual-board architecture separates the MOSFET power stage from the control-and-sensing board, preserving compact packaging while improving signal isolation across both actuator families.
The layout thermally isolates control electronics from the MOSFET bridge.

\rph{Chassis Conductive Cooling}
Heat from the stator and MOSFET bridge is passively conducted into the actuator chassis through thermal pads and the motor housing, as shown in Fig.~\ref{fig:heat_path}.
The actuator chassis and attached robot-arm mechanism therefore form a shared passive heat path for continuous-current operation without dedicated airflow.
Section~\ref{subsec:thermal_analysis} identifies and validates this thermal path.

\section{Results and Experiments}
\label{section:03_results-experiments}

\subsection{Upper-Arm Actuator Analysis}
\label{subsec:upper_arm_actuator_analysis}

\rph{Overview}
This experimental analysis verifies that the 3K compound planetary transmission satisfies the torque-transparency and packaging requirements in Sections~\ref{subsec:task_derived_parameters} and~\ref{subsec:actuator_design}.

\subsubsection{Backlash-Friction Tradeoff}
\rph{Experiment setup}
A momentum-wheel experiment characterizes the relationship between transmission backlash and static friction.
Low-amplitude torque perturbations reveal input-side backlash while the output is held quasi-static by a large-inertia wheel; the torque required to initiate motion at the same position provides the friction estimate.
Measurements are repeated throughout loaded running-in.

\begin{figure}[!!th]
    \centering
    \includegraphics[width=\columnwidth]{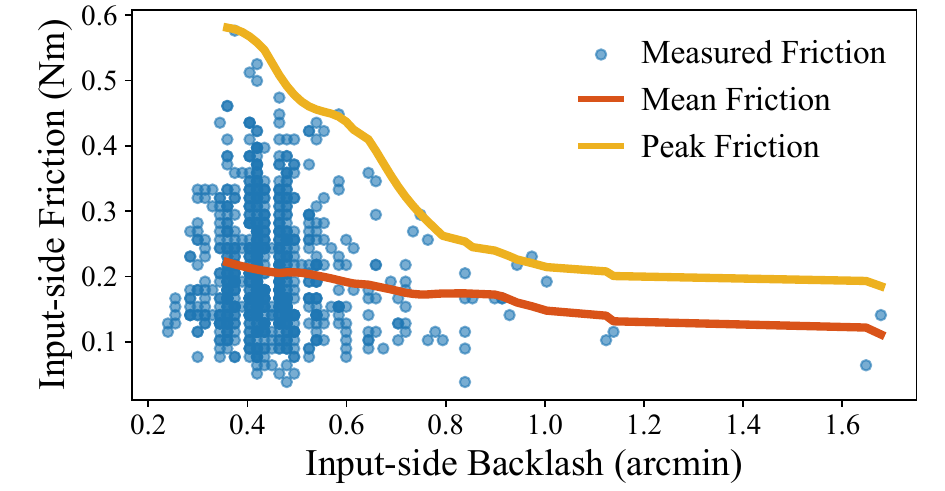}
    \vspace{-20pt}
    \caption{Backlash-friction tradeoff of the upper-arm actuator.
    Minimum driving torque required to initiate motion is plotted against input-side backlash throughout a running-in process.
    Experimental samples and corresponding mean and peak friction trends show that static friction decreases as backlash increases, indicating strain-induced resistance in pre-loaded gear meshes. 
    }
    \vspace{-5pt}
    \label{fig:backlash_friction}
\end{figure}

\begin{figure}[!t]
    \centering
    \includegraphics[width=\columnwidth]{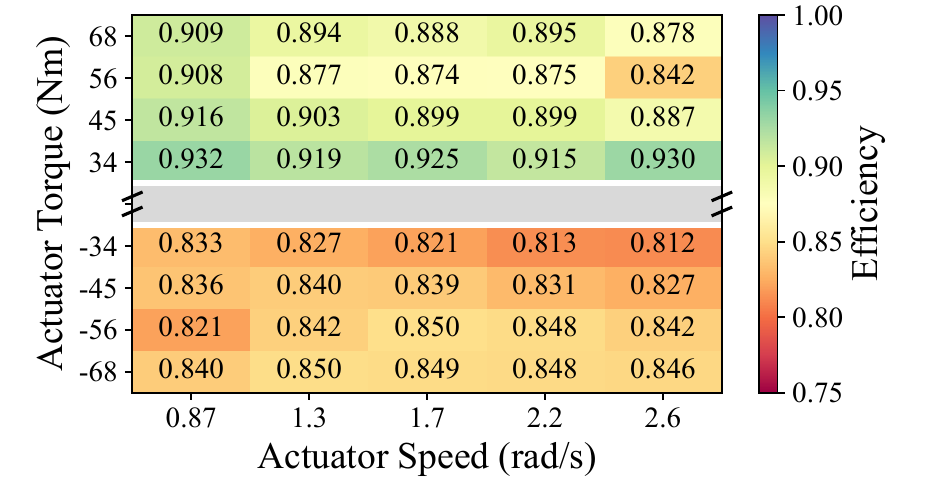}
    \vspace{-20pt}
    \caption{Upper-arm actuator bilateral efficiency.
    The upper-arm actuator achieves an average efficiency of approximately 90\% in forward driving and 84\% in backward driving over the evaluated operating range.
    }

  \label{fig:bilateral_efficiency}
\end{figure}

\rph{Preload-Induced Friction Coupling}
Static friction decreases while input-side backlash increases during running-in, as shown in Fig.~\ref{fig:backlash_friction}.
This inverse relationship suggests that a significant portion of static friction originates from elastic strain and contact deformation in preloaded gear meshes rather than simple Coulomb friction.

\subsubsection{Bilateral Efficiency}

\rph{Experiment setup}
A dynamometer couples two identical upper-arm actuators through a torque sensor, with one applying torque against the other's regulated load speed.

\rph{Efficiency Result}
Fig.~\ref{fig:bilateral_efficiency} shows the measured forward- and backward-driving efficiencies of the upper-arm actuator over the relevant operating range.
Forward efficiency remains between \(0.88\) and \(0.93\), while the lower backward efficiency is consistent with directional efficiency asymmetry in geared transmissions \cite{Bilateral_Drive_Gear}.

\subsection{Endurance Bound Analysis}
\label{subsec:thermal_analysis}

\subsubsection{Thermal Identification}
\begin{figure}[!t]
    \centering
    \captionsetup[subfloat]{position=bottom,farskip=2pt,nearskip=2pt}

    \vspace{-10pt}
    \subfloat[Transient temperature response of upper-arm actuators (Axes 1--4).\label{subfig:thermal_validation_a}]{
        \includegraphics[width=\linewidth]{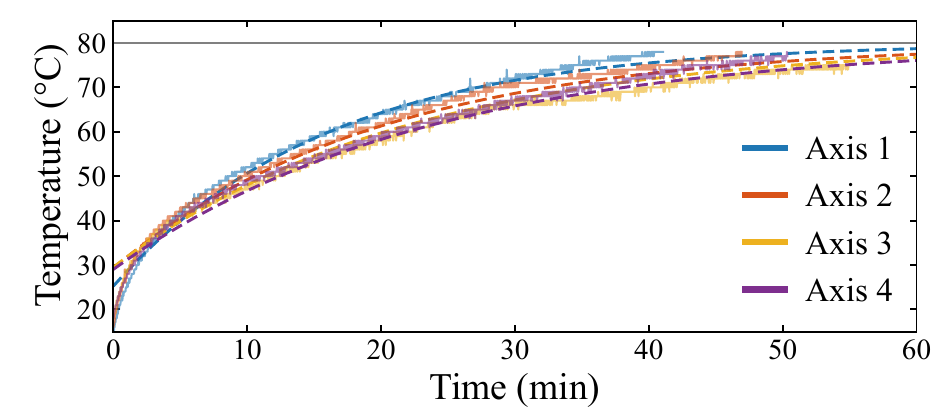}
    }
    
    \vspace{-5pt}
    
    \subfloat[Transient temperature response of forearm actuators (Axes 5--7).\label{subfig:thermal_validation_b}]{
        \includegraphics[width=\linewidth]{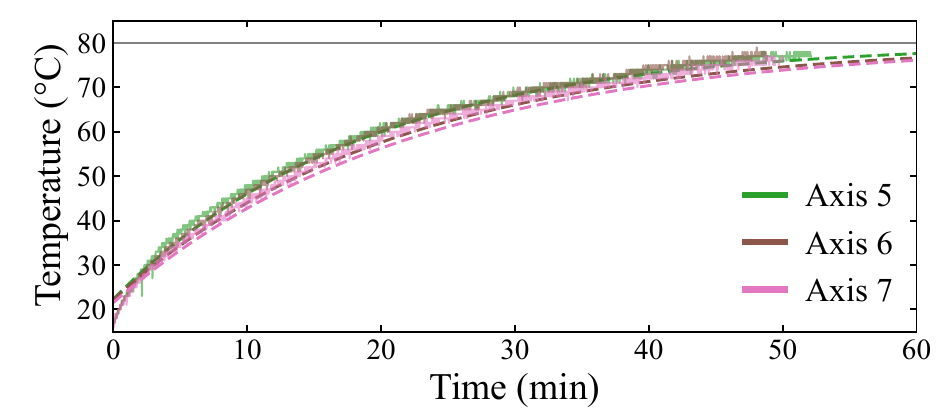}
    }

    \caption{Axis-wise transient temperature response.
    Solid curves show measured actuator temperatures; dotted curves show model predictions.
    Each joint was operated at its predicted continuous torque rating, and the measured temperatures were recorded over time. }
    \vspace{-3pt}
    \label{fig:thermal_analysis_validation}
\end{figure}

As described in Section~\ref{subsec:human_task_characteristics}, sustained joint operation of humanoids is bounded by thermal limits.
To identify the robot's thermal SOA, each joint was driven with constant current at \(20~^\circ\mathrm{C}\) ambient temperature.
The transient response was fit with a first-order model:
\begin{align}
    T(t) = T_{\mathrm{amb}} + P R \left(1 - e^{-t/(RC)}\right),
    \label{eq:thermal_rc_model}
\end{align}
where \(T(t)\) is actuator temperature, \(T_{\mathrm{amb}}\) is ambient temperature, \(P\) is total electrical loss, and \(R\) and \(C\) are the identified thermal resistance and capacitance \cite{RAIBO2}.
The continuous power limit is computed as \(P_{\mathrm{allow}}=(T_{\max}-T_{\mathrm{amb}})/R\), with continuous torque obtained from the actuator loss model.
Table~\ref{tab:thermal_analysis_result} summarizes the identified parameters and continuous limits for each axis.
Fig.~\ref{fig:thermal_analysis_validation} shows close agreement between measured and predicted temperatures, validating the model.

\subsubsection{Endurance Envelope}
\begin{figure}[!t]
    \centering
    \vspace{-15pt}
    \includegraphics[width=\columnwidth]{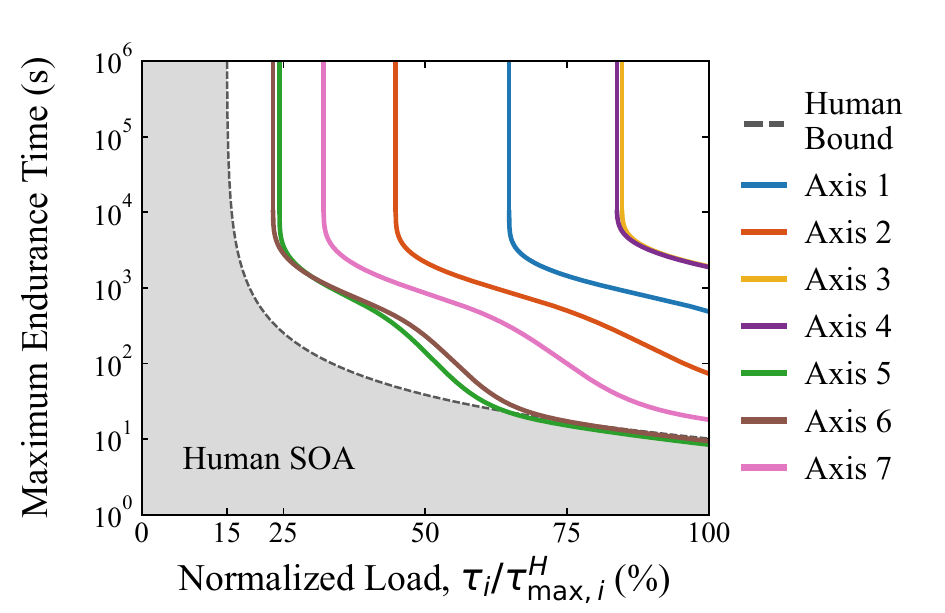}
    \vspace{-15pt}
    \caption{Endurance envelope comparison in joint-wise endurance-load space.
    The gray region denotes the human SOA, and the regions below the colored boundaries denote the axis-wise robot thermal SOA.
    {}
    }
    \label{fig:thermal_envelope_verification}
\end{figure}

\rph{Human-Task Endurance Reference}
To evaluate the
task-derived endurance target,
the robot thermal SOA and the human endurance boundary are compared in a common joint-wise endurance--load space.
Each joint load is mapped to a normalized human load coordinate as \(\tau_i/\tau_{\max,i}^{H}\), where \(\tau_{\max,i}^{H}\) denotes the corresponding source-derived, axis-mapped human torque reference defined in Section~\ref{subsec:task_derived_parameters}.
Rohmert's maximum-endurance-time relation {maps normalized human static load} to endurance-time, with loads below 15\% treated as continuously sustainable \cite{rohmert1973problems}.
Similarly, the robot thermal model predicts continuous operation below the normalized continuous-torque limit. Both thresholds appear as vertical asymptotes in Fig.~\ref{fig:thermal_envelope_verification}.

\begin{table}[!t]
    \vspace{-5pt}
    \caption{Axis-wise thermal identification and continuous operating limits of the integrated robot arm}
    \label{tab:thermal_analysis_result}
    \centering
    \footnotesize
    \setlength{\tabcolsep}{3.6pt}
    \renewcommand{\arraystretch}{1.1}
    \begin{tabular}{@{}c l c c c c@{}}
        \hline
        Axis & Type & $R$ & $C$ & $P_{\mathrm{allow}}$ & Cont. torque \\
             &      & (K/W) & (J/K) & (W) & (Nm) \\
        \hline
        1 & Upper-arm actuator & 1.09 & 762.87 & 43.46 & 64.49 \\
        2 & Upper-arm actuator & 1.55 & 672.22 & 28.35 & 48.07 \\
        3 & Upper-arm actuator & 1.58 & 719.26 & 27.60 & 47.11 \\
        4 & Upper-arm actuator & 1.26 & 965.42 & 35.00 & 55.90 \\
        5 & Forearm actuator   & 2.54 & 378.85 & 19.29 & 3.56  \\
        6 & Forearm actuator   & 2.96 & 364.94 & 16.66 & 3.19  \\
        7 & Forearm actuator   & 2.88 & 392.97 & 17.30 & 3.28  \\
        \hline
    \end{tabular}
    
    \vspace{2pt}
    \parbox{\linewidth}{\footnotesize \textit{Note:} \(P_{\mathrm{allow}}\) assumes \(T_{\max}=80~^\circ\mathrm{C}\); continuous torques are obtained from the actuator loss model. Axis-wise thermal parameters differ because attached mechanisms provide additional heat paths.}

\end{table}

\begin{figure}[!t]
    \centering
    \includegraphics[width=\linewidth]{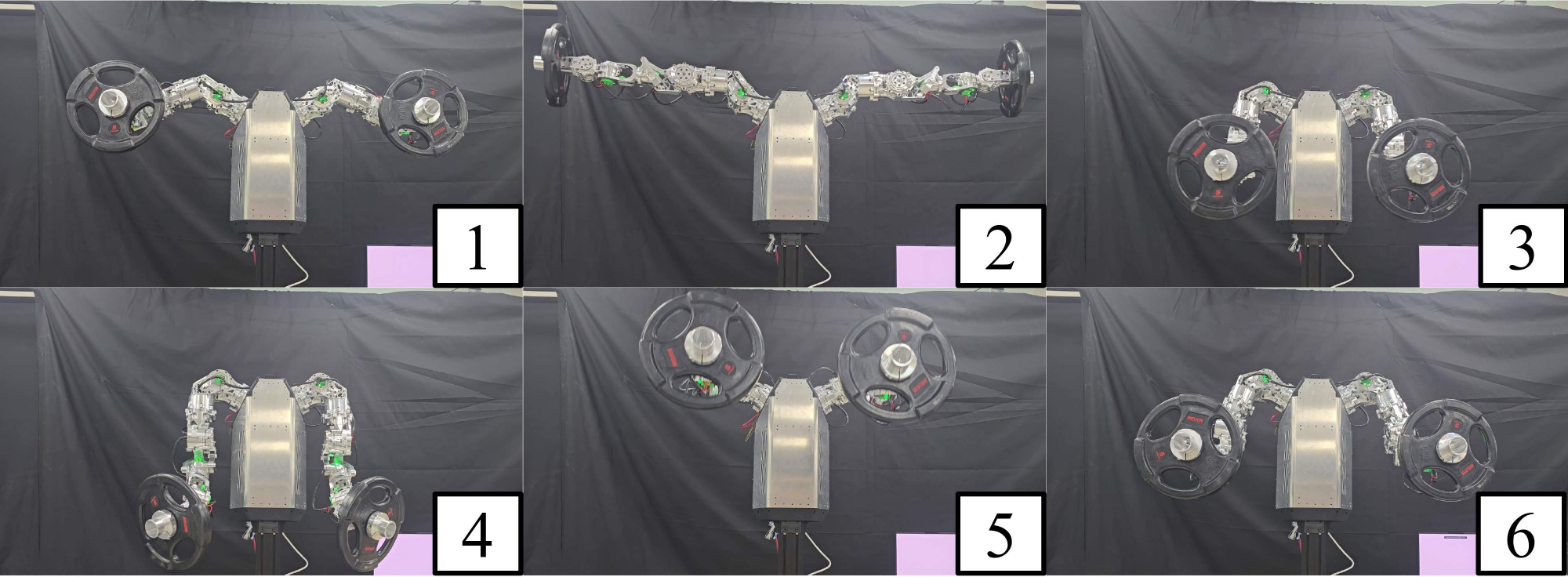}
    \vspace{-15pt}
    \caption{Continuous payload demonstration sequence.
    The robot manipulates two 5~kg discs, one per arm, along a trajectory including worst-case poses for each joint, continuously for 25~minutes.}
    \vspace{-5pt}

    \label{fig:continuous_payload_demo}
\end{figure}

\rph{Conductive Regime}
At low and moderate loads, sustainable duration is governed by the RC model because heat can diffuse from the stator and power stage into the actuator chassis and arm structure.
Conduction sets the robot's steady-state thermal boundary.

\rph{Quasi-Adiabatic Regime}
At high loads, Joule heating rises faster than conductive spreading to the chassis, so short-duration overloads are bounded by a quasi-adiabatic relation:
\begin{align}
    \left({\tau_i}/{\tau_{\max,i}^{R}}\right)^2 t_{\mathrm{sus}} = \mathrm{const.},
    \label{eq:quasi_adiabatic_limit}
\end{align}
where \(\tau_{\max,i}^{R}\) is the robot peak torque of axis \(i\), and \(t_{\mathrm{sus}}\) is the sustainable duration at the applied joint torque.
For each axis, the final robot boundary in Fig.~\ref{fig:thermal_envelope_verification} is obtained by geometrically blending the conductive and quasi-adiabatic limits with a logistic weight constrained by empirical data.

\rph{Endurance Verification}
As shown in Fig.~\ref{fig:thermal_envelope_verification}, the robot thermal SOA {largely} encloses the human endurance envelope for all axes.
Thus, the robot exceeds the human endurance envelope across {the vast majority of the} evaluated {load range}.

\vspace{-7pt}
\subsection{Demonstrations}
\label{subsec:demonstrations}

\rph{Overview}
{Four} demonstrations evaluate whether the integrated arm can realize the broad task envelope targeted by the
human-task-derived engineering parameters,
spanning force, speed, and high-impact tasks.
{Each reported demonstration was repeated up to ten times, reproducing the reported qualitative outcome or quantitative result without maintenance between trials or observable damage to the hardware components.}
All experiments used manually designed joint-space trajectories tracked by a joint-level impedance controller running at \(15~\mathrm{kHz}\), with trajectory targets updated at \(2~\mathrm{kHz}\) and fixed PD gains used across all demonstrations.

\subsubsection{Continuous Payload}
\rph{Nominal load handling}
In the continuous-payload demonstration shown in Fig.~\ref{fig:continuous_payload_demo}, the robot manipulated 5~kg per arm for 25~minutes along a trajectory {passing through the worst-case pose for} each joint, while all actuator temperatures remained below the prescribed limit.
This provides task-level evidence of repeated, thermally safe manipulation rather than a standalone endurance measure.

\begin{figure}[!htb]
    \centering
    \captionsetup[subfloat]{position=bottom,farskip=2pt,nearskip=2pt}
    
    \subfloat[Peak payload demonstration sequence.\label{subfig:demo_peak_payload_motion}]{%
        \includegraphics[width=\linewidth]{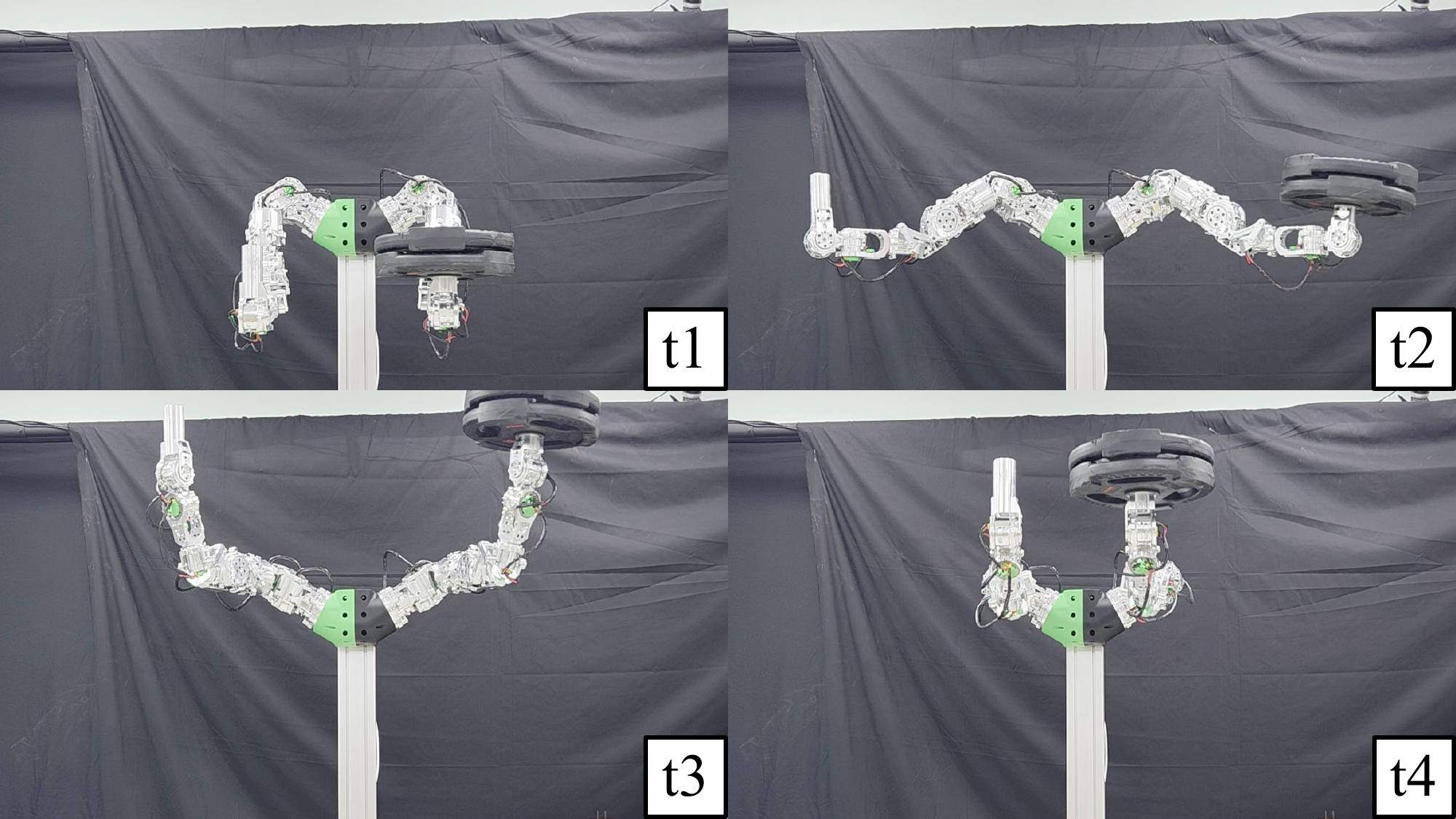}%
    }
    
    \subfloat[Upper-arm torque data during peak payload demonstration.\label{subfig:demo_peak_payload_torque}]{%
        \includegraphics[width=\linewidth]{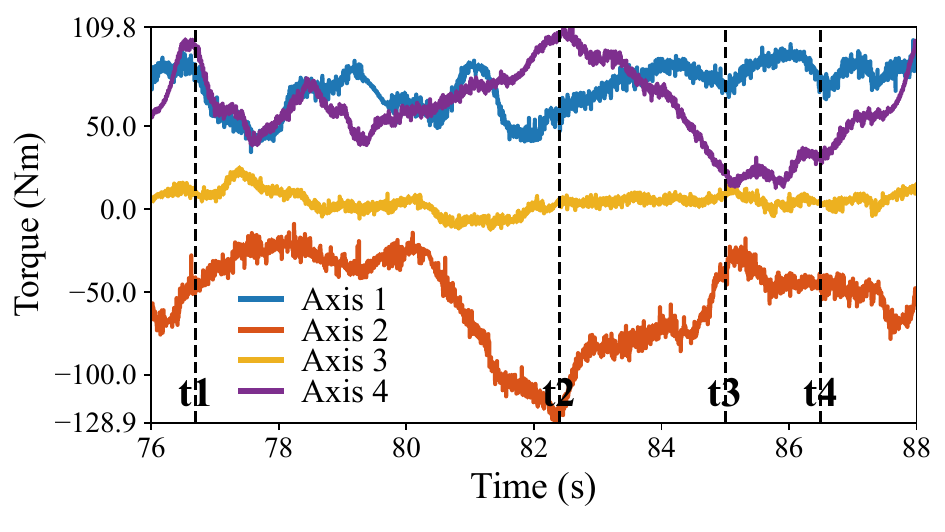}%
    }

    \caption{Peak payload demonstration.
    The robot manipulates two 10~kg discs with one hand for 60~seconds.
    The trajectory resembles human lifting with neutral forearm positions, while also including an outward extension to probe a workspace wider than the typical lifting motion.}
    \vspace{1pt}
    \label{fig:peak_payload_demo}
\end{figure}

\vspace{2.0625\baselineskip}
\subsubsection{Peak Payload}
\rph{Peak force validation}
To evaluate short-burst force capability, the robot repeatedly manipulated a 20~kg payload with one arm for 60~seconds as shown in Fig.~\ref{subfig:demo_peak_payload_motion}.
The trajectory included the extended posture \(t2\), which is more demanding than the contracted postures typically used in human lifting and therefore provides a conservative verification of peak force capability.
Fig.~\ref{subfig:demo_peak_payload_torque} confirms that the upper-arm actuators operate in their high-torque regions, particularly at \(t2\), despite the lower-torque forearm axes.
This experiment validates coordinated high-force operation across the robot task envelope.
{In particular, the Axis~4 elbow-torque trace exceeds \(100~\mathrm{Nm}\), showing that a fixed low reduction selected only to match the elbow speed used during pitching in Section~\ref{subsec:demonstrations}.\ref{subsubsec:pitching_demo} would be insufficient for the demonstrated high-payload motion.}

\vspace{1.0625\baselineskip}
\subsubsection{Pitching}
\label{subsubsec:pitching_demo}
\begin{figure}[htb]
    \centering
    \captionsetup[subfloat]{position=bottom,farskip=2pt,nearskip=2pt}
    
    \subfloat[Pitching demonstration sequence.\label{subfig:demo_pitching_main}]{%
        \includegraphics[width=\linewidth]{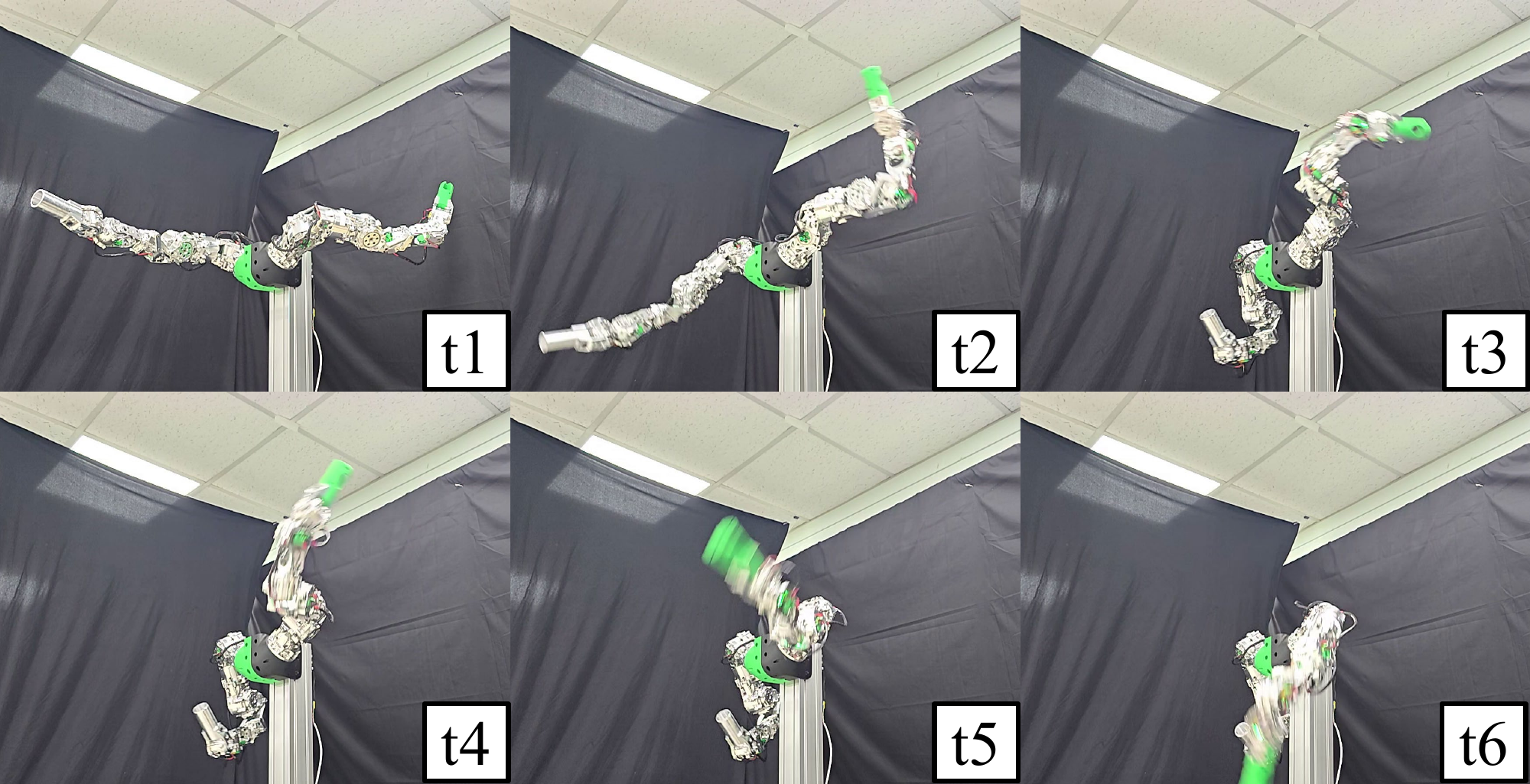}%
    }
    
    \vspace{-0.1em}
    
    \subfloat[Endpoint speed and axis 6 (forearm flexion) angular speed.\label{subfig:demo_pitching_speed}]{%
        \includegraphics[width=\linewidth]{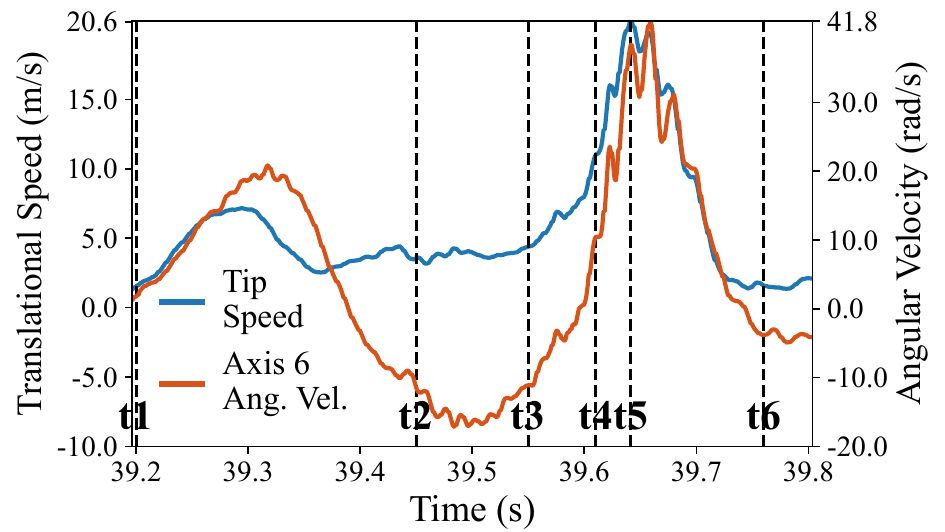}%
    }
    
    \vspace{-0.1em}
    
    \subfloat[Elbow operating regions (parallelograms) and pitching\protect\\ torque--speed trace (points) at representative times.\label{subfig:demo_pitching_or}]{%
        \includegraphics[width=\linewidth]{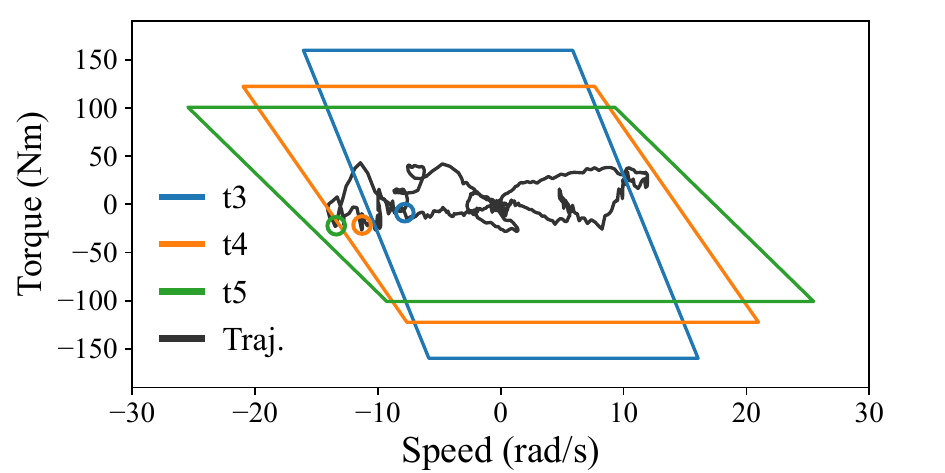}%
    }
    
    \caption{Pitching demonstration. The fast forearm and posture-adjusted elbow operating region produce a peak endpoint speed of 20.6~m/s.}
    \vspace{1pt}
    \label{fig:demo_pitching_motion}
\end{figure}

\rph{Whole-arm speed generation}
The pitching demonstration reached a peak endpoint speed of \(20.6\,\mathrm{m/s}\), approximately \(130\%\) of the average human reference.
Fig.~\ref{subfig:demo_pitching_speed} confirms rapid distal acceleration near the peak-velocity point \(t5\).
Furthermore, torque-speed analysis in Fig.~\ref{subfig:demo_pitching_or} shows that the elbow operating region expands speed-wise as the elbow extends into the peak-velocity posture \(t5\).
These measurements show that the endpoint-speed result arises from the intended directive for distal acceleration and posture-dependent speed expansion rather than an isolated actuator peak.
This verifies coordinated athletic motion through posture-dependent elbow action.

\begin{figure}[htb]
    \centering
    \captionsetup[subfloat]{position=bottom,farskip=2pt,nearskip=2pt}

    \subfloat[Hammering demonstration sequence.\label{subfig:demo_hammering_sequence}]{
        \includegraphics[width=0.95\linewidth]{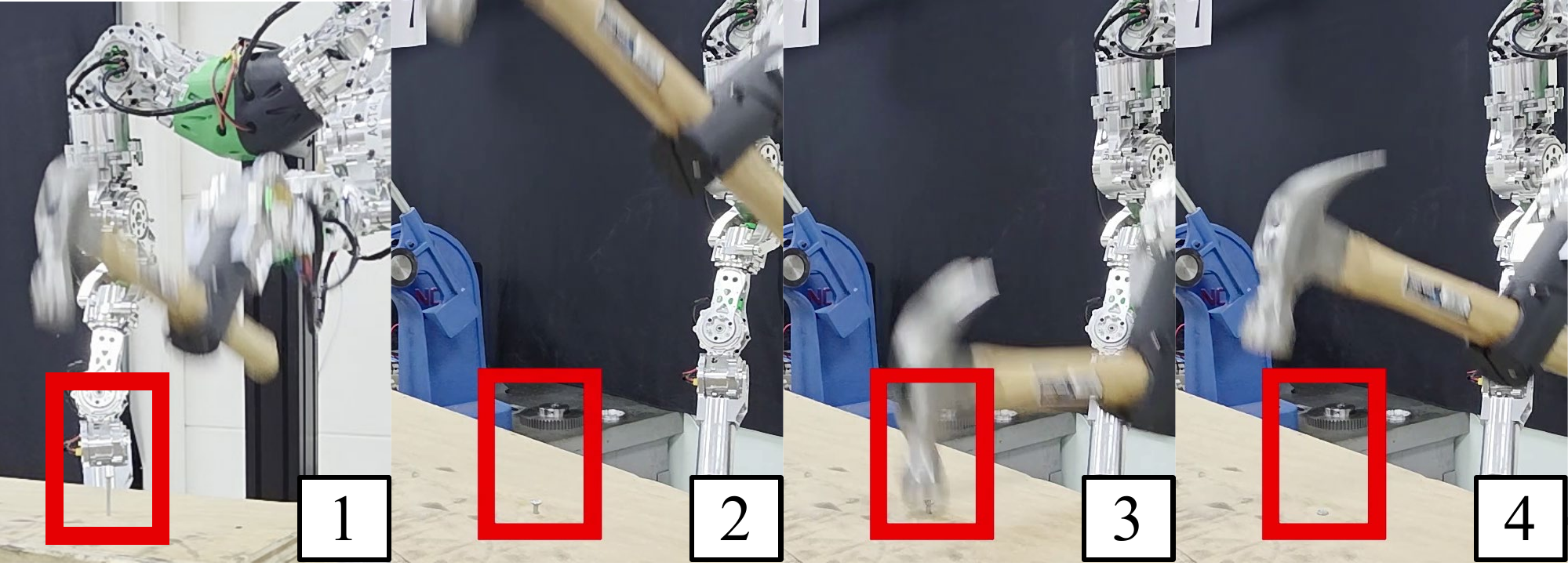}
    }

    \vspace{0.2em}

    \subfloat[{Robot-only whole-left-arm (hammer excluded)} kinetic energy over three instrumented strikes.\label{subfig:demo_hammering_energy}]{
        \includegraphics[width=0.95\linewidth]{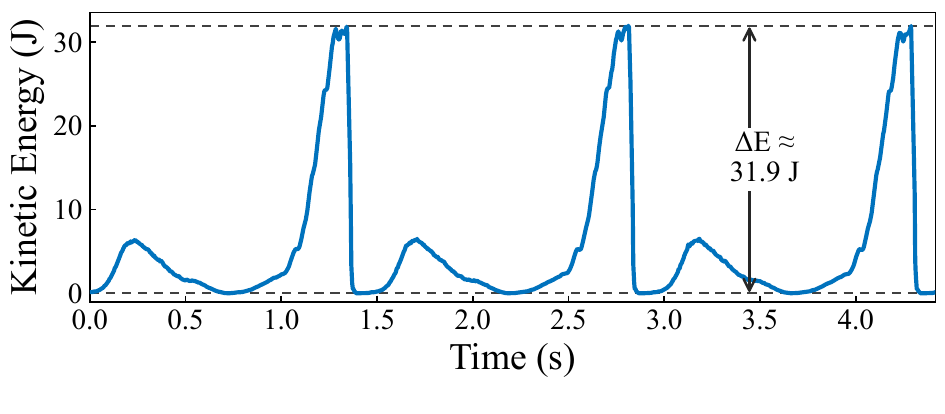}
    }

    \caption{Hammering demonstration sequence.
    The robot drives {ten pre-positioned nails} with seven strikes {each (70 total), without maintenance or observable hardware damage}. The red box highlights the tacked nail.
    Three instrumented strikes show approximately \(31.9~\mathrm{J}\) of {robot-only} arm kinetic energy{ (hammer excluded)} dissipated at impact.
`    }
    \label{fig:demo_hammering}
\end{figure}

\newpage
\subsubsection{Hammering}
\rph{Impact robustness}
The hammering demonstration shown in Fig.~\ref{fig:demo_hammering} evaluates whether the {complete arm can sustain repeated impact-rich tool use and continue operating} without {maintenance or} observable structural damage.
The task {uses} constant control gains{}. {Seven consecutive strikes per nail were repeated on ten nails (70 total impacts) without intervening maintenance or observable hardware damage.}
{The hammer assembly had a mass of $0.753~\mathrm{kg}$ and an overall length of $0.312~\mathrm{m}$. Its average impact velocity over the 70 strikes was $3.8~\mathrm{m/s}$, with an average estimated hammer contact impulse of $14.6~\mathrm{N\,s}$.}
The {robot-only complete-left-arm} kinetic energy was logged throughout three {representative} strikes{, with the hammer mass and inertia excluded}; Fig.~\ref{subfig:demo_hammering_energy} shows approximately \(32~\mathrm{J}\) of {robot-side} kinetic energy dissipated nearly instantaneously at each impact. {}
{The repeated energy dissipation through impacts relevant to human tasks demonstrates human-level capability in contact-rich settings.}

\newpage
\section{Conclusion}
\label{section:04_conclusion}
\rph{Paper Summary}
This paper presented a humanoid manipulator built for human tasks
by deriving engineering targets from published task characteristics and measurements
rather than
imposing uniform workspace-wide sizing.
This process yielded a force-specialized upper-arm, a speed-specialized forearm, a posture-dependent elbow mechanism, heterogeneous proximal and distal actuators, and compact actuator-embedded electronics with chassis conductive cooling.
Experiments verified the resulting hardware through a 20~kg peak payload, 20.6~m/s endpoint speed, endurance exceeding the human endurance envelope {across nearly the entire load range}, and high-impact {}demonstrations.
Together, these results show that the developed arm combines force, speed, endurance, and physical interaction capabilities within a single integrated humanoid manipulator.
\rph{Discussion}
The results suggest that humanoid-arm capability is shaped not only by actuator technology, but also by hardware requirement choices.
By using
human-task-derived engineering targets,
the proposed approach distributes capability across specialized arm regions instead of oversizing every joint for uniform workspace guarantees.
This provides a practical route toward humanoid manipulators that cover a broader human-task envelope without being forced into specialized regimes or devising complex energy-storing mechanisms.

\ph{Future Work}
A natural next step is to integrate the arm into a full humanoid and explore challenging whole-body locomotion and manipulation tasks.
Its low reflected inertia, broad task capability, and torque-transparent actuation also motivate {more sophisticated control methods for a broader range of human tasks}.

\newpage
\renewcommand{\IEEEbibitemsep}{1.85pt}
\bibliographystyle{IEEEtran}
\bibliography{manuscript/ref}

@IEEEtranBSTCTL{IEEEexample:BSTcontrol,
  CTLuse_forced_etal       = "yes",
  CTLmax_names_forced_etal = "6",
  CTLnames_show_etal       = "1",
  CTLdash_repeated_names   = "no"
}

@article{Faity2021TorqueReserve,
  author  = {Faity, Germain and Mottet, Denis and Pla, Simon and Froger, J{\'e}r{\^o}me},
  title   = {The Reserve of Joint Torque Determines Movement Coordination},
  journal = {Sci. Rep.},
  year    = {2021},
  volume  = {11},
  number  = {1},
  pages   = {23008},
  doi     = {10.1038/s41598-021-02338-4}
}

@ARTICLE{Bilateral_Drive_Gear,
  author={Matsuki, Hiroshi and Nagano, Kenta and Fujimoto, Yasutaka},
  journal={IEEE/ASME Trans. Mechatronics},
  title={Bilateral Drive Gear—A Highly Backdrivable Reduction Gearbox for Robotic Actuators}, 
  year={2019},
  volume={24},
  number={6},
  pages={2661--2673},
  doi={10.1109/TMECH.2019.2946403}}

@ARTICLE{Lims-ambidex,
  author={Kim, Yong-Jae},
  journal={IEEE Trans. Robot.},
  title={Anthropomorphic Low-Inertia High-Stiffness Manipulator for High-Speed Safe Interaction}, 
  year={2017},
  volume={33},
  number={6},
  pages={1358--1374},
  doi={10.1109/TRO.2017.2732354}}

@misc{ARMADA,
      title={A low-cost and lightweight 6 DoF bimanual arm for dynamic and contact-rich manipulation}, 
      author={Jaehyung Kim and Jiho Kim and Dongryung Lee and Yujin Jang and Beomjoon Kim},
      year={2025},
      note={arXiv:2502.16908},
}

@book{Sciavicco2000,
  author    = {Lorenzo Sciavicco and Bruno Siciliano},
  title     = {Modelling and Control of Robot Manipulators},
  edition   = {2},
  publisher = {Springer London},
  year      = {2000},
  doi       = {10.1007/978-1-4471-0449-0}
}

@book{Winter2009,
  author    = {David A. Winter},
  title     = {Biomechanics and Motor Control of Human Movement},
  edition   = {4},
  publisher = {John Wiley \& Sons},
  address   = {Hoboken, NJ},
  year      = {2009},
  isbn      = {9780470398180},
  doi       = {10.1002/9780470549148}
}

@article{An1981,
  author    = {K. N. An and F. C. Hui and B. F. Morrey and R. L. Linscheid and E. Y. S. Chao},
  title     = {Muscles across the elbow joint: A biomechanical analysis},
  journal   = {J. Biomech.},
  volume    = {14},
  number    = {10},
  pages     = {659--669},
  year      = {1981},
  doi       = {10.1016/0021-9290(81)90048-8}
}

@misc{RAIAthenaZero2026,
  author       = {{Robotics and AI Institute}},
  title        = {{AthenaZero}: A Bimanual Robot for Dynamic Manipulation},
  howpublished = {Blog post},
  year         = {2026},
  month        = apr,
  url          = {https://rai-inst.com/resources/blog/},
  note         = {Accessed: 2026-07-14}
}

@INPROCEEDINGS{D3-ARM,
  author={Luo, Hong and Xu, Jianle and Li, Shoujie and Liang, Huayue and Chen, Yanbo and Xia, Chongkun and Wang, Xueqian},
  booktitle={Proc. IEEE Int. Conf. Robot. Autom.},
  title={D3-ARM: High-Dynamic, Dexterous and Fully Decoupled Cable-Driven Robotic Arm}, 
  year={2025},
  volume={},
  number={},
  pages={6252--6258},
  doi={10.1109/ICRA55743.2025.11127687}}

@INPROCEEDINGS{IIT_Bi-Manual,
  author={Baccelliere, Lorenzo and Kashiri, Navvab and Muratore, Luca and Laurenzi, Arturo and Kamedula, Małgorzata and Margan, Alessio and Cordasco, Stefano and Malzahn, Jörn and Tsagarakis, Nikos G.},
  booktitle={Proc. IEEE/RSJ Int. Conf. Intell. Robots Syst.},
  title={Development of a human size and strength compliant bi-manual platform for realistic heavy manipulation tasks}, 
  year={2017},
  volume={},
  number={},
  pages={5594--5601},
  doi={10.1109/IROS.2017.8206447}}

@misc{DuaticDynaArm,
  author       = {{Duatic AG}},
  title        = {{DynaArm}: Ultra-Lightweight Robotic Arm},
  howpublished = {\url{https://www.duatic.com/dynaarm}},
  note         = {Accessed: 2026-07-14}
}

@article{RAIBO2,
  author = {Choongin Lee and Donghoon Youm and Jeongsoo Park and Jeonghyun Lee and
            Suyoung Choi and Gwanghyeon Ji and Juhyeok Mun and Moonkyu Jung and
            Hyungho Chris Choi and Hyeongjun Kim and Hyunsik Oh and Kyeongmin Nam and
            Minho Lee and Jungwoo Hur and Donghyuk Choi and Donggeon Kim and
            Yeonsu An and Jemin Hwangbo},
  title   = {RAIBO2: Highly efficient quadruped robot completing full marathon with a single battery charge},
  journal = {Research Square},
  year    = {2025},
  doi     = {10.21203/rs.3.rs-6040970/v1},
  note    = {Preprint, Mar. 26, 2025, doi: 10.21203/rs.3.rs-6040970/v1}
}

@article{rohmert1973problems,
  title={Problems in determining rest allowances: Part 1: use of modern methods to evaluate stress and strain in static muscular work},
  author={Rohmert, Walter},
  journal={Appl. Ergon.},
  volume={4},
  number={2},
  pages={91--95},
  year={1973},
  publisher={Elsevier}
}

@article{Hirashima2008KineticChain,
  author  = {Hirashima, Masaya and Yamane, Katsu and Nakamura, Yoshihiko and Ohtsuki, Tatsuyuki},
  title   = {Kinetic chain of overarm throwing in terms of joint rotations revealed by induced acceleration analysis},
  journal = {J. Biomech.},
  volume  = {41},
  number  = {13},
  pages   = {2874--2883},
  year    = {2008},
  doi     = {10.1016/j.jbiomech.2008.06.014}
}

@article{Lategan2011IsokineticNorms,
  author  = {Lategan, Leon},
  title   = {Isokinetic norms for ankle, knee, shoulder and forearm muscles in young {South African} men},
  journal = {Isokinet. Exerc. Sci.},
  volume  = {19},
  number  = {1},
  pages   = {23--32},
  year    = {2011},
  doi     = {10.3233/IES-2011-0392}
}

@article{LateganKruger2007Handgrip,
  author  = {Lategan, Leon and Kr{\"u}ger, P. E.},
  title   = {Effect of two different handgrip positions on elbow peak torque values},
  journal = {S. Afr. J. Res. Sport Phys. Educ. Recreation},
  volume  = {29},
  number  = {2},
  pages   = {67--74},
  year    = {2007},
  doi     = {10.4314/sajrs.v29i2.25969}
}

@article{Delp1996WristMoments,
  author  = {Delp, Scott L. and Grierson, Andrea E. and Buchanan, Thomas S.},
  title   = {Maximum isometric moments generated by the wrist muscles in flexion-extension and radial-ulnar deviation},
  journal = {J. Biomech.},
  volume  = {29},
  number  = {10},
  pages   = {1371--1375},
  year    = {1996},
  doi     = {10.1016/0021-9290(96)00029-2}
}

@article{JessopPain2016MaximumVelocities,
  author  = {Jessop, David M. and Pain, Matthew T. G.},
  title   = {Maximum velocities in flexion and extension actions for sport},
  journal = {J. Hum. Kinet.},
  volume  = {50},
  pages   = {37--44},
  year    = {2016},
  doi     = {10.1515/hukin-2015-0139}
}

@incollection{KeyserlingArmstrong2008Anthropometry,
  author    = {Keyserling, W. Monroe and Armstrong, Thomas J.},
  title     = {Ergonomics and Work-Related Musculoskeletal Disorders},
  booktitle = {Wallace/Maxcy-Rosenau-Last Public Health \& Preventive Medicine},
  edition   = {15},
  editor    = {Wallace, Robert B.},
  publisher = {McGraw-Hill},
  address   = {New York, NY, USA},
  pages     = {763--779},
  year      = {2008}
}

@inproceedings{tomishiro2019design,
  title={Design of Robot Leg with Variable Reduction Ratio Crossed Four-bar Linkage Mechanism},
  author={Tomishiro, Kohei and Sato, Ryuki and Harada, Yuji and Ming, Aiguo and Meng, Fei and Liu, Huaxin and Fan, Xin and Chen, Xuechao and Yu, Zhangguo and Huang, Qiang},
  booktitle={Proc. IEEE/RSJ Int. Conf. Intell. Robots Syst.},
  pages={4333--4338},
  year={2019},
}

@article{lenzi2011robotic,
  title={A robotic model to investigate human motor control},
  author={Lenzi, Tommaso and Vitiello, Nicola and McIntyre, Joseph and Roccella, Stefano and Carrozza, Maria Chiara},
  journal={Biol. Cybern.},
  volume={105},
  number={1},
  pages={1--19},
  year={2011},
  publisher={Springer}
}

@article{Holzbaur2007UpperLimb,
  author  = {Holzbaur, Katherine R. S. and Murray, Wendy M. and
             Gold, Garry E. and Delp, Scott L.},
  title   = {Upper Limb Muscle Volumes in Adult Subjects},
  journal = {J. Biomech.},
  volume  = {40},
  number  = {4},
  pages   = {742--749},
  year    = {2007},
  doi     = {10.1016/j.jbiomech.2006.11.011}
}

@INPROCEEDINGS{KUKA_DLR_LWR,
  author={Bischoff, Rainer and Kurth, Johannes and Schreiber, Guenter and Koeppe, Ralf and Albu-Schaeffer, Alin and Beyer, Alexander and Eiberger, Oliver and Haddadin, Sami and Stemmer, Andreas and Grunwald, Gerhard and Hirzinger, Gerhard},
  booktitle={Proc. 41st Int. Symp. Robot. and 6th German Conf. Robot.},
  title={The KUKA-DLR Lightweight Robot arm - a new reference platform for robotics research and manufacturing}, 
  year={2010},
  volume={},
  number={},
  pages={1--8},
  doi={}}
\end{document}